# Investigating Large Language Models in Inferring Personality Traits from User Conversations

Jianfeng Zhu[1], Ruoming Jin [1], and Karin G. Coifman[2]
1 Department of Computer Science, Kent State University, Kent, OH 44224
2 Department of Psychological Sciences, Kent State University, Kent, OH 44224
*Jianfeng Zhu.
**Email:** jzhu10@kent.edu

## Abstract

Large Language Models (LLMs) are demonstrating remarkable human-like capabilities across diverse domains, including psychological assessment. This study evaluates whether LLMs, specifically GPT-4o and GPT-4o mini, to infer Big Five personality traits and generate Big Five Inventory-10 (BFI-10) item scores from user conversations under zero-shot prompting conditions. Our findings reveal that incorporating an intermediate step—prompting for BFI-10 item scores before calculating traits—enhances accuracy and aligns more closely with the gold standard than direct trait inference. This structured approach underscores the importance of leveraging psychological frameworks in improving predictive precision. Additionally, a group comparison based on depressive symptom presence revealed differential model performance. Participants were categorized into two groups: those experiencing at least one depressive symptom and those without symptoms. GPT-4o mini demonstrated heightened sensitivity to depression-related shifts in traits such as Neuroticism and Conscientiousness with the symptom-present group, whereas GPT-4o exhibited strengths in nuanced interpretation across groups. These findings underscore the potential of LLMs to analyze real-world psychological data effectively, offering a valuable foundation for interdisciplinary research at the intersection of artificial intelligence and psychology.

## Significance Statement

This study advances the emerging field of AI-driven personality assessment by showcasing the potential of LLMs to complement traditional psychometric tools. By integrating structured frameworks, the findings underscore the sensitivity of LLMs to psychological states and their ability to detect nuanced differences in mental health markers. These contributions enhance theoretical understanding of AI applications in psychological contexts, while paving the way for future research and practical innovations in mental health diagnostics and interventions.

## Main Text

### Introduction

Mental health is a critical component of overall well-being and remains at the forefront of global health challenges [1]. Personality traits, as delineated by the Costa and McCrae five-factor model, consist of Extraversion, Agreeableness, Conscientiousness, Neuroticism, and Openness, and have been shown to significantly influence mental health outcomes [2]. Meta-analyses have emphasized the importance of personality traits in understanding individual differences in psychopathology [3-4] and well-being [5-6]. For instance, personality disorders are often viewed as extreme expressions of personality traits such as high neuroticism and low agreeableness [7]. Moreover, maladaptive traits exacerbate stress, a key variable in mental health, leading to negative downstream effects on physical health and behaviors [8-9]. For example, low agreeableness has been associated with heightened stress levels, adversely affecting biological markers of physical health [10] and contribute to maladaptive coping behaviors [11]. As such, personality differences often manifest earlier and more directly in mental health outcomes compared to their effects on physical health or health-related behaviors.

Recent advancements in artificial intelligence, particularly in deep learning and Large Language Models (LLMs), have opened new avenues for predicting Big Five personality traits. These models provide innovative approaches for understanding personality through textual data, presenting alternatives to traditional self-report questionnaires. LLMs, such as GPT, have demonstrated their capabilities in simulating and predicting personality traits using advanced natural language processing techniques. For example, a recent study used role-playing simulations and questionnaire – based prompting to enhance prediction accuracy for Big Five personality traits from counseling dialogues [12]. Other studies have investigated prompt engineering techniques to predict traits like extraversion and assess their impact on model accuracy and robustness [13-14]. Research also indicates that larger parameter sets in LLMs exhibit nuanced personality traits, such as increased openness and conscientiousness, while fine-tuned models display minor modulations in specific traits [15].

Despite these advancements, in the use of Large Language Models (LLMs) for personality inference, most existing approaches [12-15] rely on directly predicting personality traits without leveraging the structure relationship between traits and widely accepted psychological tools like the Big Five Inventory (BFI) [16]. The core principle of the Big Five model is rooted in linguistic encoding, as represented by tools such as the Big Five Inventory-44 (BFI-44) or its condensed forms like the BFI-10 [17-18]. This raises a crucial question: Can LLMs effectively utilize BFI questions to infer personality traits? Moreover, can LLMs accurately predict individual BFI item scores, which serve as a foundation for calculating these traits? To the best of knowledge, existing research has not explored these questions using semi-structured ideographic interview conversations.

Given the intrinsic connection between personality and mental wellness, it is also essential to ask: **Can LLMs predict individual BFI item scores and how accurately they are? And how does personality prediction based on these predicted BFI scores compared with directly using LLM for personality trait prediction? In addition, since personality traits are inherently related to mental health, does LLM prediction for BFI and personality differentiate with respect to mental health conditions?** Investigating these questions can provide valuable insights into the capacity of LLMs for identifying and differentiating markers of mental health, paving the way for applications that enhance human mental wellness and improve the characterization of **LLM-generated personas** [19].

This study seeks to bridge the gap by examining the capacity of LLMs to deduce Big Five personality traits using two distinct approaches: (1) directly prompting the LLM to infer Big Five traits from user-generated text, and (2) prompting the LLM to produce BFI-10 item scores, which are then used to compute Big Five traits using established scoring guidelines. Comparing these approaches allows for an assessment of the accuracy, alignment, and potential advantages of integrating structured psychological frameworks into LLM-based personality inference.

To achieve these objectives, this study employs a unique dataset comprising responses from 102 participants engaged in semi-structured interviews. These participants were responded to five standardized questions covering topics such as their activities from the previous days, a recent challenging experience, coping mechanisms, an unpleasant event, and a positive occurrence. Additionally, participant's responses were paired with clinical depression symptom labels derived from Structured Clinical Interviews for DSM-5 (SCID-5) administered by trained clinicians [20], as well as ground-truth Big Five personality trait scores measured using the BFI-10.

Understanding the strengths and limitations of LLMs in inferring sensitive psychological traits from real-world data is paramount importance, especially as their applications in research and practical contexts continue to grow. While LLMs provide unprecedented opportunities for gaining insights into users' psychological profiles, their capability raises ethical concerns, particularly around the potential to craft personalized, persuasive messages tailored to individual's personality traits. However, the lack of sufficient validation of these methods amplifies ethical issues regarding their use both in research and in industry. This dual-edged nature underscores the need for new forms of AI governance and regulation to address these emerging challenges.

In this study, we examine whether LLMs can infer psychological traits without explicit task-specific training, utilizing a zero-shot learning paradigm. Specifically, we employ OpenAI's ChatGPT (GPT-4o and GPT-4o mini) [21] to infer Big Five personality traits and BFI-10 item scores based on data from semi-structured interview conversations. Building on prior research that used direct prompts to infer Big Five traits, we further investigate whether leveraging BFI-10 items scores can enhance prediction accuracy.

Our research is guided by the following key research questions:

**RQ1:** How do LLMs such as GPT-4o and GPT-4o mini differ in their ability to infer Big Five traits from user conversations using prompted responses?

**RQ2:** How accurately can LLMs predict BFI-10 item scores?

**RQ3:** How does the direct prediction of Big Five traits by GPT models compare to their calculation utilizing LLM-prompted BFI-10 scores?

**RQ4:** How well do LLMs capture differences in Big Five personality traits between users with and without experiencing depressive symptoms?

**RQ5:** How effectively do LLM models reflect differences in BFI-10 item scores between users with and without experiencing depressive symptoms?

## Materials and Methods

### Dataset

The dataset for this study comprises previously collected speech responses from 102 participants. Of these. Demographic information indicate that 96 participants identified as male (94%) and 6 identified as female (6%). Participants' age ranged from 24 to 56 years, with a mean age of 36.63 years and a total age range of 32 years. The data were obtained through recorded, semi-structured, ideographic interview sessions.

During these interviews, all participants answered the same five questions in a predetermined sequence. The first question required participants to describe their activities from waking to sleep on the previous day. The subsequent questions explored their personal experiences with a recent challenging event or situation, their coping strategies for that challenge, an unrelated recent unpleasant event, and, finally, a recent positive experience.

Interviewers were trained to provide standardized prompts, and participants were encouraged to speak spontaneously for up to three minutes in response to each question. Each interview session resulted in approximately 15 minutes of recorded speech per participant. The speech was then recorded and transcribed according to convention and recommendations [22-25].

Each participant's text responses were paired with their corresponding depressive symptoms, derived from Structured Clinical Interview for DSM (SCID) reports. Additionally, ground truth personality trait scores, measured using the Big Five Inventory-10 (BFI-10), were included.

For preprocessing, text responses were standardized by converting all text to lowercase, removing common English stopwords, and filtering out punctuation. These steps ensured consistency and enhanced the analytical reliability of the data.

### Methods

### Applying the Big Five Personality Framework to LLMs

This study utilizes the Costa and McCrae five-factor model of personality [16] to evaluate the Big Five traits (Extraversion, Agreeableness, Conscientiousness, Neuroticism, and Openness). Prompts were designed to instruct the language models to assign scores on a Likert scale ranging from 1 to 5 to participant responses, referencing the specific facets associated with each Big Five personality domain (Table 1).

Table 1: Trait facets associated with the five domains of the Costa and McCrae five-factor model of personality [26]

| Conscientiousness | Order, dutifulness, achievement striving, self-discipline, deliberation |
| Agreeableness | Trust, straightforwardness, altruism, compliance, modesty, tender-mindedness |

| Neuroticism | Anxiety, angry hostility, depression, self-consciousness, impulsiveness, vulnerability |
|---|---|
| Openness | Fantasy, aesthetics, values |
| extraversion | Warmth, gregariousness, assertiveness, excitement-seeking |

The models were prompted using the following instructions:

> "You are an expert psychologist specializing in personality analysis.
> Based on the Big Five Personality Traits model, you will evaluate an individual's responses to five questions.
> Each response reflects different dimensions of personality traits.
> For each of the Big Five traits, consider the following facets.
>   Conscientiousness: order, dutifulness, achievement striving, self-discipline, deliberation.
>   Agreeableness: trust, straightforwardness, altruism, compliance, modesty, tendermindedness.
>   Neuroticism: anxiety, angry hostility, depression, self-consciousness, impulsiveness, vulnerability.
>   Openness: fantasy, aesthetics, values.
>   Extraversion: warmth, gregariousness, assertiveness, excitement-seeking.
>  Instructions:
>      Read the individual's responses to five questions carefully.
>      For each personality trait, assign a score between 1 (Very Low) and 5 (Very High) based on the themes, tone, and content of the responses.
>      Each score can be rounded to the nearest tenth, like 3.5.
>      Scoring Guide:
>      1: Very Low - The response shows little to no alignment with the trait's facets.
>      2: Low - The response shows weak alignment with the trait's facets.
>      3: Moderate - The response shows some alignment but not strongly.
>      4: High - The response strongly aligns with the trait's facets.
>      5: Very High - The response shows exceptional alignment with the trait's facets.
>      {questions_text}
>      Conversation: {conversation}
>      ### Task:
>   Analyze the given text for Big Five personality traits and provide the scores in the following format without any explanation or extra words:
>     "Conscientiousness": score, "Agreeableness": score, "Neuroticism": score, "Openness": score, "Extraversion": score
>                  "

**Incorporating BFI-10 Items for LLM Scoring**

A separate prompt was used to generate scores for the BFI-10 framework, which consists of 10 items representing the Big Five traits. Each item was rated on a scale of 1 (Strongly Disagree) to 5 (Strongly Agree) according to the participant's responses. The following instructions were provided:

> """
>    You are a psychologist trained in analyzing personality traits using the Big Five Inventory (BFI-10).
>    Your task is to score the responses to each BFI-10 question on a scale from 1 (Strongly Disagree) to 5 (Strongly Agree),
>    provide a detailed explanation for the score, and reference specific parts of the input text.
>
>    Below is a transcript from a semi-structured psychological interview for the five questions:
>     {questions_text}
>      Instructions:
>      Read the individual's responses to five questions carefully.

> For each personality trait, assign a score between 1 (Very Low) and 5 (Very High) based on the themes, tone, and content of the responses.
> Scoring Guide:
> 1: Very Low - The response shows little to no alignment with the trait's facets.
> 2: Low - The response shows weak alignment with the trait's facets.
> 3: Moderate - The response shows some alignment but not strongly.
> 4: High - The response strongly aligns with the trait's facets.
> 5: Very High - The response shows exceptional alignment with the trait's facets.
> The BFI-10 framework items are:
> 1. (BFI_1) Is reserved
> 2. (BFI_2) Is generally trusting
> 3. (BFI_3) Tends to be lazy
> 4. (BFI_4) Is relaxed, handles stress well
> 5. (BFI_5) Has few artistic interests
> 6. (BFI_6) Is outgoing, sociable
> 7. (BFI_7) Tends to find fault with others
> 8. (BFI_8) Does a thorough job
> 9. (BFI_9) Gets nervous easily
> 10. (BFI_10) Has an active imagination
> Here is the full transcript from the five questions:
> Conversation: {conversation}
> ### Task:
> Analyze the given text for Big Five Inventory and provide the scores in the following format without any explanation and extra word:
> "BFI_1" :Score, "BFI_2":Score, ..., "BFI_10": Score .
> """

The generated BFI-10 scores were subsequently used to compute the Big Five trait scores. These scores were compared with GPT-inferred scores to assess their alignment with the golden standard benchmarks.

**LLM Sensitivity to Personality Variations Linked to Depressive Symptoms**

To classify participants, a survey like the PHQ-9 was employed, categorizing individuals into two groups: "Depressive Symptom Present" (1), for participants reporting at least one symptom, and "No Depressive Symptom" (0), for those reporting none. While this approach does not constitute a formal clinical diagnosis, it provides a reliable means of identifying mental health-related group differences.

This analysis examines the sensitivity of Large Language Models (LLMs) in detecting personality differences associated with depressive symptoms. Differences in Big Five personality traits were analyzed between the two groups using both GPT-inferred personality predictions from participant text responses and BFI-10-derived personality scores, which served as the golden standard. The sensitivity of GPT models was evaluated by analyzing the magnitude and direction of personality trait differences between the groups, comparing these findings to those observed in BFI-10-derived scores. This approach aims to determine whether LLM-based methods can reliably detect mental health-related personality differences, highlighting their potential applications in mental health research.

**Performance Evaluation Approach**

The performance of GPT-4o and GPT-4o mini was evaluated using two metrics: correlation and mean difference. Correlation measured the ability to capture the structural relationships within personality traits, while mean difference quantified absolute prediction accuracy by measuring deviations from the gold standard scores.

Although both metrics provide valuable insights, our analysis primarily focused on mean differences, as they directly reflect the models' accuracy in replicating ground-truth personality trait and BFI-10 items scores. This emphasis is particularly relevant for practical applications, such as mental health research or personalized interventions, which precise trait predictions are essential. By prioritizing mean difference, we ensure a rigorous evaluation of how closely the models' predictions align with validated psychological data.

## Results
**RQ1: How do LLMs such as GPT-4o and GPT-4o mini differ in their ability to infer Big Five traits from user conversations using prompted responses?**

The ability of GPT-4o and GPT-4o mini to infer personality traits was assessed by comparing their performance against the golden standard using correlation coefficients and mean differences for each trait: extraversion, agreeableness, conscientiousness, neuroticism, and openness (Table 1, Figure 1).

Table 1 Correlation and Mean Difference Score for GPT-4 and GPT-4 Mini.

| Big Five Trait | Corr_Golden_vs_GPT4o | MeanDiff_Golden_vs_GPT4o | Corr_Golden_vs_GPT4oMini | MeanDiff_Golden_vs_GPT4oMini |
| --- | --- | --- | --- | --- |
| Extraversion | 0.098 | **-0.514** | 0.151 | -0.578 |
| Agreeableness | 0.184 | **-0.147** | -0.063 | -0.338 |
| Conscientiousness | -0.058 | 0.353 | -0.019 | **0.157** |
| Neuroticism | 0.025 | -0.985 | -0.09 | **-0.848** |
| Openness | 0.142 | 0.382 | 0.185 | **0.162** |

For extraversion, both models demonstrated weak positive correlation (0.098 for GPT-4o; 0,151 for GPT-4O mini), with GPT-4o mini slightly outperforming GPT-4o. However, both models underestimated extraversion, as evidenced by negative mean differences ( -0.514 for GPT-4o and -0.578 for GPT-4o mini). For agreeableness, GPT-4o exhibited a weak positive correlation (0.184) and smaller mean difference (-0.147), outperforming GPT-4o mini, which showed a slight negative correlation (-0.063) and a larger mean difference (-0.338). For conscientiousness, both models showed weak negative correlations with the golden standard (-0.058 for GPT-4o and -0.019 for GPT-4o mini). GPT-4o produced a mean difference of 0.353, while GPT-4o mini showed a smaller mean difference of 0.157. In predicting neuroticism, GPT-4o had a near-zero positive correlation (0.025) and a large negative mean difference (-0.985), while GPT-4o mini had a weak negative correlation (-0.090) and a smaller mean difference (-0.848). For openness, GPT-4o Mini slightly outperformed GPT-4o with higher correlations (0.185 vs. 0.142) and smaller mean differences (0.162 vs. 0.382).
.

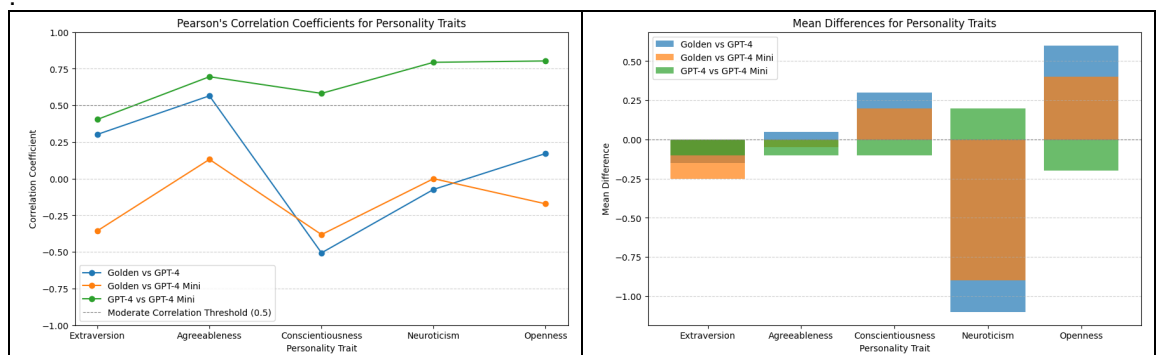

Figure 1 Person's Correlation Coefficients and Mean Differences for Personality Traits

**RQ2: How accurately can LLMs predict BFI-10 item scores?**

The Table 2 and Figure 2 a nuanced performance across individual BFI-10 items. GPT-4o excelled in predicting certain items such as BFIS_1 (1.058 vs. 1.108 for GPT-4o mini), BFIS_3 (-0.196 vs. -0.608), BFIS_5 (0.265 vs. 0.824), and BFIS_9 (-1.137 vs. -1.186). While GPT-4o mini showed superior performance in other items like BFIS_2 (0.176 vs. 0.333 for GPT-4o), BFIS_4 (0.225 vs. 0.569), BFIS_6 (-0.029 vs. 0.725), BFIS_7 (-0.020 vs. 0.225), BFIS_8 (0.343 vs. 0.402), and BFIS_10 (0.235 vs. 0.441). These findings suggest that GPT-4o mini may have an advantage in structured psychometric tasks, particularly for items requiring nuanced interpretation, whereas GPT-4o performs better in specific domains with less variability.

Table 2 Comparison of Mean Differences Between GPT-4o and GPT-4o Mini Across BFI Items

| BFI_Item | Corr_Original_vs_GPT4o | MeanDiff_Original_vs_GPT4o | Corr_Original_vs_GPT4o Mini | MeanDiff_Original_vs_GPT4o Mini |
|---|---|---|---|---|
| BFIS_1 | 0.152796 | **1.058824** | 0.071523 | 1.107843 |
| BFIS_2 | 0.255887 | 0.333333 | -0.025557 | **0.176471** |
| BFIS_3 | 0.035116 | **-0.196078** | 0.018442 | -0.607843 |
| BFIS_4 | 0.078101 | 0.568627 | -0.216461 | **0.22549** |
| BFIS_5 | -0.108095 | **0.264706** | 0.021945 | 0.823529 |
| BFIS_6 | 0.238155 | 0.72549 | 0.171492 | **-0.029412** |
| BFIS_7 | 0.259168 | 0.22549 | 0.079641 | **-0.019608** |
| BFIS_8 | 0.033102 | 0.401961 | -0.001926 | **0.343137** |
| BFIS_9 | 0.00632 | **-1.137255** | -0.02882 | -1.186275 |
| BFIS_10 | -0.047342 | 0.441176 | 0.056938 | **0.235294** |

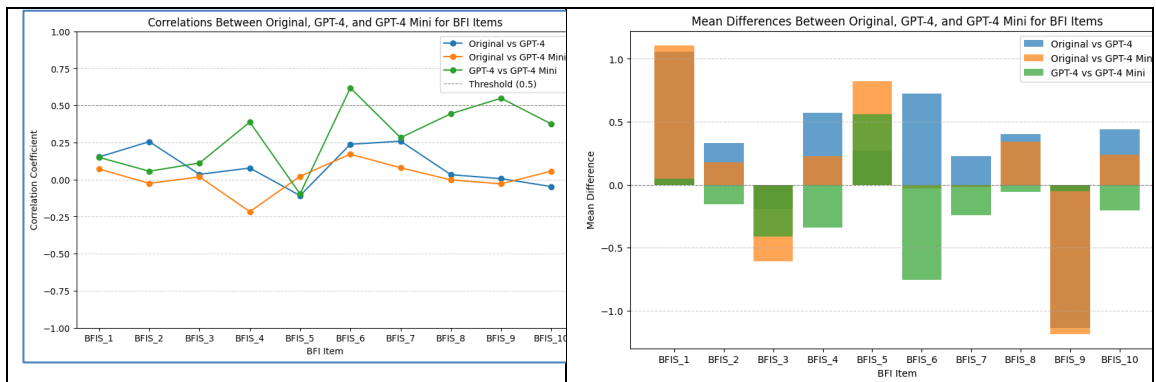

Figure 2 Person's Correlation Coefficients and Mean Differences for BFI 10

**RQ3: How does the direct prediction of Big Five traits by GPT models compare to their calculation utilizing LLM-prompted BFI-10 scores?**

Table 3 and Figure 3 summarize the differences between direct prediction of Big Five traits by GPT models and their calculation using LLM-prompted BFI-10-based score. The findings reveal that utilizing structured frameworks, such as BFI-10, improves alignment with the gold standard.

For extraversion, the BFI-10-based calculations significantly improved accuracy for both GPT-4o and GPT-4o mini, reducing mean differences to 0.598 and 0.245, respectively, compared to direct predictions (-0.514 and -0.578). Similarly for agreeableness, BFI-10-based scores for GPT-4o (mean difference: 0.054) and GPT-4o mini (0.098) were substantially closer to the gold standard than direct predictions (-0.147 for GPT-4o and -0.338 for GPT-4o mini). For conscientiousness, both models showed moderate improvements with BFI-10 scores, yielding mean difference of 0.299 for GPT-4o and 0.475 for GPT-4o mini.

In the case of neuroticism, although BFI-10-based calculations also improved accuracy, the changes were less substantial, with GPT-4o's mean difference improving from -0.985 to -0.853 and GPT-4o mini's improving from -0.848 to -0.706. Lastly, for openness, BFI-10 scores significantly enhanced model alignment with the gold standard. GPT-4o's mean difference improved to 0.088 (from 0.382), highlighting the positive impact of structured frameworks on prediction accuracy.

Table 3 Comparison of Mean Differences in Big Five Personality Traits Scores Using LLM-Based BFI-10 Calculations Versus Direct Predictions for GPT-4o and GPT-4o mini

| Trait | MeanDiff_Golden_vs_GPT4o | MeanDiff_Golden_vs_GPT4oBFI | MeanDiff_Golden_vs_GPT4o mini | MeanDiff_Golden_vs_GPT4o miniBFI |
|---|---|---|---|---|
| Extraversion | -0.514 | 0.598039 | -0.578 | **0.245098** |
| Agreeableness | -0.147 | **0.053922** | -0.338 | 0.098039 |
| Conscientiousness | 0.353 | 0.29902 | 0.157 | 0.47549 |
| Neuroticism | -0.985 | -0.852941 | -0.848 | **-0.705882** |
| Openness | 0.382 | 0.088235 | 0.162 | -0.294118 |

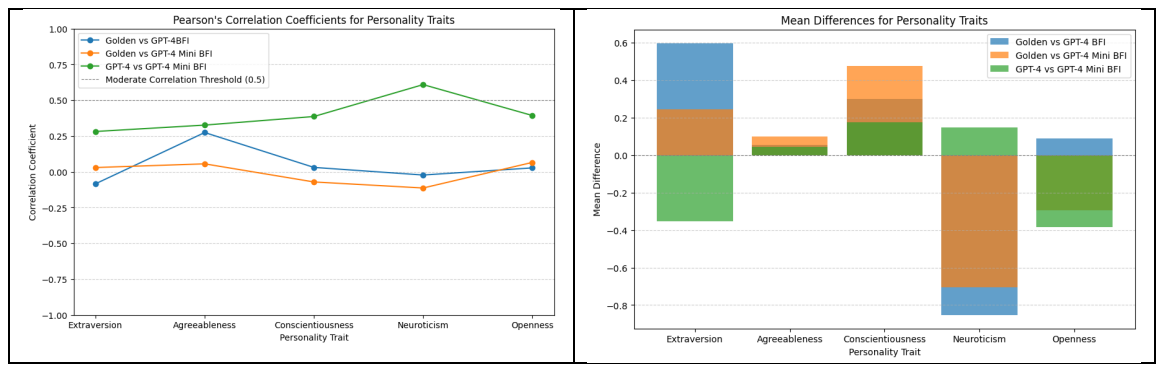

Figure 3 Person's Correlation Coefficients and Mean Differences for Personality Trait of BFI 10

**RQ4: How well do LLMs capture differences in Big Five personality traits between users with and without experiencing depressive symptoms?**

An analysis of trait predictions across symptom groups (Table 4, Figure 4), revealed distinct performance trends for the models. In the symptom-present group, both GPT-4o and GPT-4o mini exhibited smaller mean differences for conscientiousness and neuroticism, indicating better alignment with the golden standard for these traits. In the symptom-absent group, agreeableness and extraversion were accurately by GPT-4o compared to GPT-4o mini, while GPT-4o mini demonstrated better performance for openness.

Table 4 Differences in Big Five Personality Trait Predictions for Participants with and without Depressive Symptoms

| Personality Traits | LLMs | Mean Difference from Golden-Without Symptoms | Difference from Golden -With Symptoms |
|---|---|---|---|
| Agreeableness | GPT-4o | **0.094937** | 0.326087 |
| Agreeableness | GPT-4o-mini | **0.28481** | 0.521739 |
| Conscientiousness | GPT-4o | -0.398734 | **-0.195652** |
| Conscientiousness | GPT-4o-mini | -0.183544 | **-0.065217** |
| Extraversion | GPT-4o | **0.512658** | 0.521739 |
| Extraversion | GPT-4o-mini | 0.594937 | **0.521739** |
| Neuroticism | GPT-4o | 1.018987 | **0.869565** |
| Neuroticism | GPT-4o-mini | 0.917722 | **0.608696** |
| Openness | GPT-4o | **-0.316456** | -0.608696 |
| Openness | GPT-4o-mini | **-0.126582** | -0.282609 |

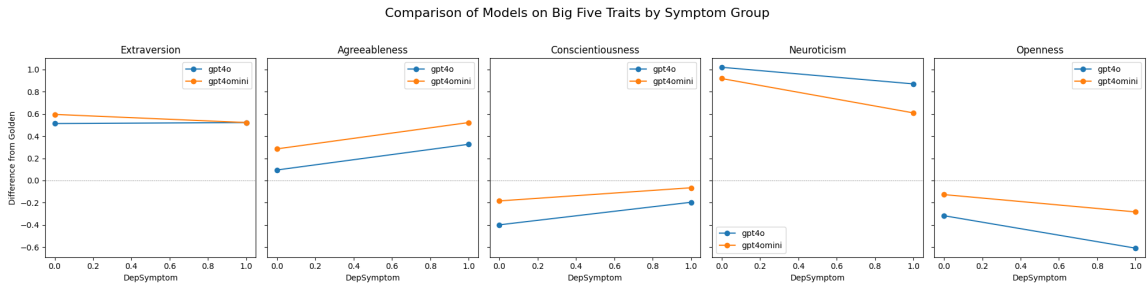

Figure 4 Comparison of GPT Models' Performance on Big Five Personality Traits Across Symptom Groups

**RQ5: How effectively do LLM models reflect differences in BFI-10 item scores between users with and without experiencing depressive symptoms?**

The analysis of BFI-10 item scores across symptom groups (Table 5, Figure 5) highlights key trends in the performance of GPT-4o and GPT-4o mini. For BFIS_1, BFIS_3, BFIS_4, BFIS_5, and BFIS_8, GPT-4o mini achieved smaller absolute mean differences in the symptom-present group, including perfect alignment for BFIS_4 (mean difference: 0.00 compared to -0.435 for GPT-4o). For BFIS_9, GPT-4o mini slightly outperformed GPT-4o, with a smaller mean difference (1.00 vs. 1.087 in the symptom-present group). Across both symptom groups, GPT-4o mini consistently outperformed GPT-4o for BFIS_2, BFIS_6, and BFIS_10, demonstrating enhanced alignment with gold standard.

Table 5 Table 5: Mean Differences in BFI-10 Item Scores from the Golden Standard Across Participants with and Without Depressive Symptoms

| BFI_10 Items | Model | Mean Difference from Golden - Without Depressive Symptoms | Mean Difference from Golden -With Depressive Symptoms |
|---|---|---|---|
| **BFI_1** | GPT-4o | -1.139241 | **-0.782609** |
| **BFI_1** | GPT-4o mini | -1.189873 | **-0.826087** |
| **BFI_2** | GPT-4o | -0.329114 | -0.347826 |
| **BFI_2** | GPT-4o mini | **-0.177215** | **-0.173913** |
| **BFI_3** | GPT-4o | 0.21519 | **0.130435** |
| **BFI_3** | GPT-4o mini | 0.658228 | **0.434783** |
| **BFI_4** | GPT-4o | -0.607595 | **-0.434783** |
| **BFI_4** | GPT-4o mini | -0.291139 | **0.0** |
| **BFI_5** | GPT-4o | -0.303797 | **-0.130435** |
| **BFI_5** | GPT-4o mini | -0.873418 | **-0.652174** |
| **BFI_6** | GPT-4o | -0.746835 | **-0.652174** |
| **BFI_6** | GPT-4o mini | **0.012658** | 0.086957 |
| **BFI_7** | GPT-4o | **-0.202532** | -0.304348 |

| BFI_7 | GPT-4o mini | **0.037975** | -0.043478 |
| BFI_8 | GPT-4o | -0.405063 | **-0.391304** |
| BFI_8 | GPT-4o mini | -0.367089 | **-0.26087** |
| BFI_9 | GPT-4o | 1.177215 | **1.0** |
| BFI_9 | GPT-4o mini | 1.21519 | **1.086957** |
| BFI_10 | GPT-4o | **-0.329114** | -0.826087 |
| BFI_10 | GPT-4o mini | **-0.151899** | -0.521739 |

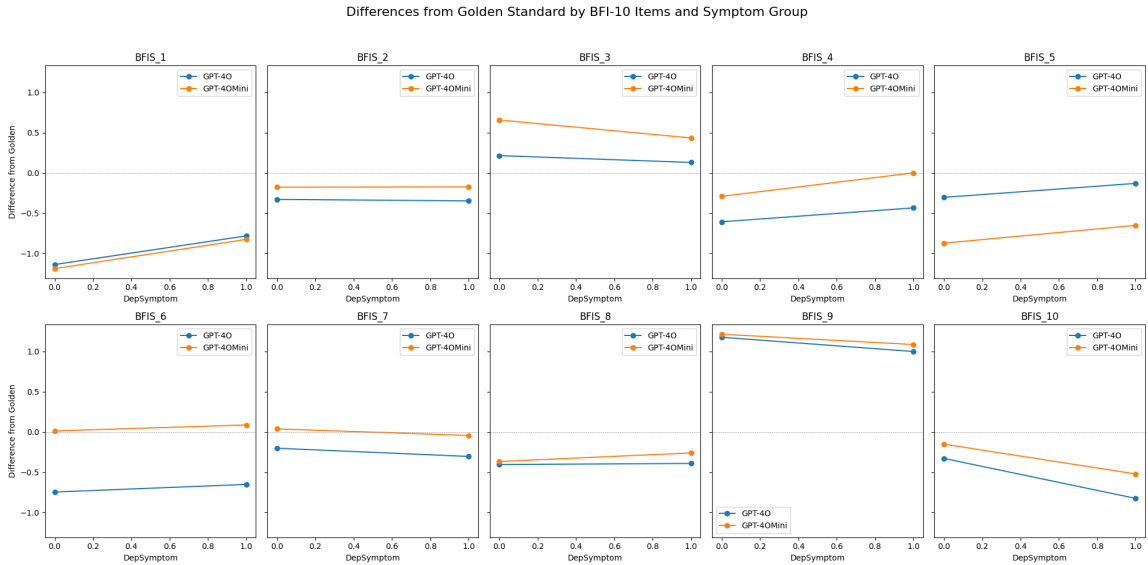

Figure 5 Difference from Golden Standard by BFI-10 Items and Symptom Group

**Discussion**
This study explored the capacity of Large Language models (LLMs), specific GPT-4o and GPT-4o mini, to infer Big Five personality traits from user conversations through prompted responses. Across five research questions, the analysis examined the nuances of personality trait inference, the effectiveness of structured frameworks like BFI-10, and the imfluence of depressive symptoms on model predictions.

**Inferring Big Five Traits from Conversations (RQ1)**
The ability of GPT-4o and GPT-4o mini to infer Big Five personality traits yielded mixed results. While correlations with the gold standard were generally weak, GPT-4o mini performed slightly better overall performance, particularly for traits such as extraversion and openness. However, both models underestimated certain traits like extraversion and neuroticism. These outcomes align with findings from PersonaLLM [27], which demonstrated that while LLMs can approximate personality traits, their accuracy varies by trait, and human evaluations of LLM outputs can differ in perception.

**Predicting BFI-10 Item Scores (RQ2)**
In predicting individual BFI-10 items scores, GPT-4o mini outperformed GPT-4o, achieving smaller mean differences across most items. This result underscores the potential of smaller, optimized models for specialized psychological tasks. Moreover, the structured design of the BFI-10 framework contributed to improved predictive accuracy, emphasizing the value of integrating psychometric tools into LLM-based assessments. These findings are supported by studies such as Predicting the Big Five Personality Traits in Chinese Counseling Dialogues [28], which showed that role-play and structured frameworks prompts enhance prediction accuracy.

**Comparison of Direct Predictions and BFI-10-Based Calculations (RQ3)**
The analysis showed that calculating Big Five traits using LLM-prompted BFI-10 item scores led to better alignment with gold-standard than direct trait predictions. This finding emphasizes the importance of structured frameworks, which provide a scaffold for improving the interpretability and accuracy of personality assessments using LLMs. Similarly, research like LLM Questionnaire Completion for Automatic Psychiatric Assessment [29] demonstrated that structured frameworks enhance diagnostic accuracy when applied to mental health contexts.

**Capturing Personality Differences in Depressive Symptom Groups (RQ4)**
LLMs demonstrated varying ability to predict personality traits between users with and without depressive symptoms. For traits like conscientiousness and neuroticism, both models aligned more closely with the gold standard in the symptom-present group. In contrast, GPT-4o generally performed better in the symptom-absent group for traits such as agreeableness and openness, while GPT-4o mini excelled at predicting extraversion in the symptom-present group. These findings resonate with insights from DepreSym [30], which examined the ability of LLMs to identify psychological markers and highlighted their capacity to differentiate between psychological states.

**Predicting BFI-10 Item Scores in Depressive Symptom Groups (RQ5)**
For BFI-10 item scores, GPT-4o mini consistently showed better performance in the symptom-present group, achieving smaller mean differences across several items. Highlights included perfect alignment for BFIS_4 (Is relaxed, handles stress well) and improved accuracy for BFIS_8 (Does a thorough job). GPT-4o mini also outperformed GPT-4o in both symptom-present and symptom-absent groups for items such as BFIS_2 (Is generally trusting) and BFIS_6 (Is outgoing, sociable). These results indicate that LLM performance can vary depending on user psychological states, pointing to their potential applications in personalized mental health assessments. Research such as DepreSym corroborates this sensitivity, showcasing the nuanced performance of LLMs in identifying depressive symptoms through structured data.

**Conclusion**

This study demonstrates the nuanced ability of large language models, particularly GPT-4o and GPT-4o mini, to infer personality traits and predict structured psychological metrics such as BFI-10 scores. The findings reveal moderate success in trait inference, with better performance observed when integrating structured psychometric frameworks. GPT-4o Mini often outperformed GPT-4o, particularly in contexts involving depressive symptoms, though both models showed areas for improvement, particularly in predicting traits like extraversion and neuroticism.

These results emphasize the importance of structured methodologies, such as BFI-10, in enhancing LLM performance and underscore the potential for using LLMs in scalable and personalized mental health assessments. However, the study's limitations—including weak correlations and limited generalizability—highlight the need for future research to refine model algorithms, validate findings across diverse populations, and explore hybrid AI-human approaches. Addressing these areas can unlock the full potential of LLMs as tools for psychological analysis and mental health innovation.

**References**

[1] Mental health. World Health Organization. Jun 17, 2022. URL: https://www.who.int/news-room/fact-sheets/detail/mental-health-strengthening-our-response [accessed 2024-12-15]
[2] Costa Jr, P. T., & McCrae, R. R. (2010). The five-factor model, five-factor theory, and interpersonal psychology. *Handbook of interpersonal psychology: Theory, research, assessment, and therapeutic interventions*, 91-104.

[3] Kotov, R., Gamez, W., Schmidt, F., & Watson, D. (2010). Linking "big" personality traits to anxiety, depressive, and substance use disorders: a meta-analysis. *Psychological bulletin*, *136*(5), 768.

[4] Malouff, J. M., Thorsteinsson, E. B., & Schutte, N. S. (2005). The relationship between the five-factor model of personality and symptoms of clinical disorders: A meta-analysis. *Journal of psychopathology and behavioral assessment*, *27*, 101-114.

[5] DeNeve, K. M., & Cooper, H. (1998). The happy personality: a meta-analysis of 137 personality traits and subjective well-being. *Psychological bulletin*, *124*(2), 197.

[6] Steel, P., Schmidt, J., & Shultz, J. (2008). Refining the relationship between personality and subjective well-being. *Psychological bulletin*, *134*(1), 138.

[7] Krueger, R. F., & Tackett, J. L. (2006). Personality and psychopathology. New York, NY: Guilford Press.

[8] Friedman, H. S. (2008). The multiple linkages of personality and disease. Brain, Behavior, and Immunity, 22, 668–675. http://dx.doi.org/10.1016/ j.bbi.2007.09.004

[9] Smith, T. W., Gallo, L. C., Shivpuri, S., & Brewer, A. L. (2012). Personality and health: Current issues and emerging perspectives. In A. Baum, T. Revenson, & J. Singer (Eds.), Handbook of health psychology (2nd ed., pp. 374–404). New York, NY: Taylor & Francis

[10] Robles, T. F., Glaser, R., & Kiecolt-Glaser, J. K. (2005). Out of balance: A new look at chronic stress, depression, and immunity. Current Directions in Psychological Science, 14, 111–115. http://dx.doi.org/10.1111/ j.0963-7214.2005.00345.x

[11] Wills, T. A., Sandy, J. M., & Yaeger, A. M. (2002). Stress and smoking in adolescence: A test of directional hypotheses. Health Psychology, 21, 122–130. http://dx.doi.org/10.1037/0278-6133.21.2.122

[12] Yan, Y., Ma, L., Li, A., Ma, J., & Lan, Z. (2024). Predicting the Big Five Personality Traits in Chinese Counselling Dialogues Using Large Language Models. *arXiv preprint arXiv:2406.17287*.

[13] Molchanova, M. (2024). Exploring the Potential of Large Language Models for Text-Based Personality Prediction. In: Rapp, A., Di Caro, L., Meziane, F., Sugumaran, V. (eds) Natural Language Processing and Information Systems. NLDB 2024. Lecture Notes in Computer Science, vol 14763. Springer, Cham. https://doi.org/10.1007/978-3-031-70242-6_28

[14] Killian, J., Sun, R. (2024). Detecting Big-5 Personality Dimensions from Text Based on Large Language Models. In: Fred, A., Hadjali, A., Gusikhin, O., Sansone, C. (eds) Deep Learning Theory and Applications. DeLTA 2024. Communications in Computer and Information Science, vol 2172. Springer, Cham. https://doi.org/10.1007/978-3-031-66705-3_18

[15] Hilliard, A., Munoz, C., Wu, Z., & Koshiyama, A. S. (2024). Eliciting Big Five Personality Traits in Large Language Models: A Textual Analysis with Classifier-Driven Approach. *arXiv preprint arXiv:2402.08341*.

[16] Costa Mastrascusa, R., de Oliveira Fenili Antunes, M. L., de Albuquerque, N. S., Virissimo, S. L., Foletto Moura, M., Vieira Marques Motta, B., ... & Quarti Irigaray, T. (2023). Evaluating the complete (44-item), short (20-item) and ultra-short (10-item) versions of the Big Five Inventory (BFI) in the Brazilian population. *Scientific Reports*, *13*(1), 7372.

[17] John, O. P. (1999). The Big-Five trait taxonomy: History, measurement, and theoretical perspectives. *Handbook of Personality: Theory and Research/Guilford*.

[18] Rammstedt, B. (2007). The 10-item big five inventory. *European Journal of Psychological Assessment*, *23*(3), 193-201.

[19] Schuller, A., Janssen, D., Blumenröther, J., Probst, T. M., Schmidt, M., & Kumar, C. (2024, May). Generating personas using LLMs and assessing their viability. In *Extended Abstracts of the CHI Conference on Human Factors in Computing Systems* (pp. 1-7).

[20] First, M. B. (2014). Structured clinical interview for the DSM (SCID). *The encyclopedia of clinical psychology*, 1-6.

[21] OpenAI. "Hello GPT-4o." *OpenAI*, OpenAI, https://openai.com/index/hello-gpt-4o/. Accessed [12/02/2024]

[22] Coifman, K. G., Bonanno, G. A., Ray, R. D., & Gross, J. J. (2007). Does repressive coping promote resilience? Affective-autonomic response discrepancy during bereavement. *Journal of personality and social psychology*, *92*(4), 745.

[23] Coifman, K.G. & Bonanno, G.A., (2010*). When distress does *not* become depression; Emotion context sensitivity and adjustment to bereavement. *Journal of Abnormal Psychology, 119(3),* 479-490*.

*[24] Harvey, M.M., Coifman, K.G., Ross, G., Kleinert, D. & Giardina, P. (2014). Contextually appropriate emotion-word use predicts adaptive health behavior: Emotion context sensitivity and treatment adherence. Journal of Health Psychology, online first: DOI: 10.1177/1359105314532152*

[25] Coifman, K.G., Flynn, J.J. & Pinto, L.A. (2016). When Context Matters: Negative emotions predict psychological health and adjustment. *Motivation & Emotion, 40(4), 602-624.*

[26] Matthews, G., Deary, I. J., & Whiteman, M. C. (2003). *Personality traits*. Cambridge University Press.

[27] Jiang, H., Zhang, X., Cao, X., Breazeal, C., Roy, D., & Kabbara, J. (2023). Personallm: Investigating the ability of large language models to express personality traits. *arXiv preprint arXiv:2305.02547*.

[28] Yan, Y., Ma, L., Li, A., Ma, J., & Lan, Z. (2024). Predicting the Big Five Personality Traits in Chinese Counselling Dialogues Using Large Language Models. *arXiv preprint arXiv:2406.17287*.

[29] Rosenman, G., Wolf, L., & Hendler, T. (2024). LLM Questionnaire Completion for Automatic Psychiatric Assessment. *arXiv preprint arXiv:2406.06636*.

[30] Pérez, A., Fernández-Pichel, M., Parapar, J., & Losada, D. E. (2023). DepreSym: A Depression Symptom Annotated Corpus and the Role of LLMs as Assessors of Psychological Markers. *arXiv preprint arXiv:2308.10758*.